\documentstyle[darpasls,times,epsfig]{article}

\begin{document}

\title{Entropy-based Pruning of Backoff Language Models}

\author{Andreas Stolcke}

\institution{Speech Technology And Research Laboratory \\
SRI International \\
Menlo Park, California}

\maketitle

\begin{abstract}
A criterion for pruning parameters from
N-gram backoff language models is developed, based on the relative entropy
between the original and the pruned model.
It is shown that the relative entropy resulting from pruning a single N-gram
can be computed exactly and efficiently for backoff models.
The relative entropy measure can be expressed as a relative
change in training set perplexity.
This leads to a simple pruning criterion whereby all N-grams 
that change perplexity by less than a threshold are removed from the 
model.
Experiments show that a production-quality Hub4 LM can be reduced to
26\% its original size without increasing recognition error.
We also compare the approach to a heuristic pruning criterion
by Seymore and Rosenfeld \cite{Seymore:96}, and show that their
approach can be interpreted as an approximation to the relative entropy
criterion.  Experimentally, both approaches select similar sets of
N-grams (about 85\% overlap), with the exact relative entropy criterion
giving marginally better performance.
\end{abstract}

\section{Introduction}

N-gram backoff models \cite{Katz:87}, despite their shortcomings,
still dominate
as the technology of choice for state-of-the-art speech recognizers
\cite{Jelinek:91c}.
Two sources of performance improvements are the use of higher-order models
(several DARPA-Hub4 sites now use 4-gram or 5-gram models) and 
the inclusion of more training data from more sources (Hub4 models
typically include Broadcast News, NABN and WSJ data).
Both of these approaches lead to model sizes that are
impractical unless some sort of parameter selection technique is
used.
In the case of N-gram models, the goal of parameter selection is to
chose which N-grams should have explicit conditional probability
estimates assigned by the model, so as to maximize performance
(i.e., minimize perplexity and/or recognition error) while minimizing
model size.
As pointed out in \cite{Kneser:96}, pruning (selecting parameters from)
a full N-gram model of higher order amounts to building a 
{\em variable-length} N-gram model, i.e., one in which training set
contexts are not uniformly represented by N-grams of the same length.

Seymore and Rosenfeld \cite{Seymore:96} showed that selecting
N-grams based on their conditional probability estimates and 
frequency of use is more effective than the traditional 
absolute frequency thresholding.
In this paper we revisit the problem of N-gram parameter selection
by deriving a criterion that satisfies the following desiderata.
\begin{itemize}
\item {\bf Soundness:} 
	The criterion should optimize some well-understood
	information-theoretic measure of language model quality.
\item {\bf Efficiency:}
	An N-gram selection algorithm should be fast, i.e.,
	take time proportional to the number of N-grams under
	consideration.
\item {\bf Self-containedness:}
	As a practical consideration, we want to be able to
	prune N-grams from existing language models.
	This means a pruning criterion should be based only on
	information contained in the model itself.
\end{itemize}

In the remainder of this paper we describe our pruning algorithm based on 
relative entropy distance between N-gram distributions
(Section~\ref{sec:pruning}),
investigate how the quantities required for the pruning criterion can
be obtained efficiently and exactly (Section~\ref{sec:computing}),
show that the criterion is highly effective in reducing the size of
state-of-the-art language models with negligible performance penalties
(Section~\ref{sec:experiments}),
investigate the relation between our pruning criterion and that
of Seymore and Rosenfeld (Section~\ref{sec:relation}),
and draw some conclusions (Section~\ref{sec:conclusions}).

\newcommand{\bow}{\alpha}
\newcommand{\isbackoff}{{\rm BO}}
\newcommand{\mysum}[1]{{\displaystyle \sum_{\makebox[2em]{\scriptsize$#1$}}}}
\newcommand{\dist}[2]{D(#1 || #2)}
\newcommand{\perpl}{{\it PP}}

\section{N-gram Pruning Based on Relative Entropy}
	\label{sec:pruning}

An N-gram language model represents a probability distribution
over words $w$, conditioned on $(N-1)$-tuples of preceding words, or
histories $h$.
Only a finite set of N-grams $(w,h)$ have conditional probabilities
explicitly represented in the model.  The remaining N-grams are assigned
a probability by the recursive backoff rule
\[
	p(w | h) = \bow(h) p(w | h')
\]
where $h'$ is the history $h$ truncated by the first word
(the one most distant from $w$), and $\bow(h)$ is a {\em backoff weight}
associated with history $h$, determined so that $\sum_{w} p(w | h) = 1$.

The goal of N-gram pruning is to remove explicit estimates
$p(w | h)$ from the model, thereby reducing the number of 
parameters, while minimizing the performance loss.
Note that after pruning, the retained explicit N-gram probabilities are
unchanged, but backoff weights will have to be recomputed, thereby 
changing the values of implicit (backed-off) probability estimates.
Thus, the pruning approach chosen is conceptually independent of the
estimator chosen to determine the explicit N-gram estimates.

Since one of our goals is to prune N-gram models without access to
any statistics not contained in the model itself, a natural criterion
is to minimize the `distance' between the distribution embodied by the
original model and that of the pruned model.
A standard measure of divergence between distributions is 
{\em relative entropy} or {\em Kullback-Leibler distance}
(see, e.g., \cite{Cover:91}).  Although not strictly a distance metric,
it is a non-negative, continuous function
that is zero if and only if the two distributions are identical.

Let $p(\cdot | \cdot)$ denote the conditional probabilities assigned by the
original model, and $p'(\cdot | \cdot)$ the probabilities in the pruned
model.
Then, the relative entropy between the two models is
\begin{equation}
\label{eq:delta}
	\dist{p}{p'} = - \sum_{w_i, h_j} p(w_i,h_j) 
				[ \log p'(w_i|h_j) - \log p(w_i | h_j) ]
\end{equation}
where the summation is over all words $w_i$ and histories (contexts) $h_j$.

Our goal will be to select N-grams for pruning such that $\dist{p}{p'}$
is minimized.
However, it would not be feasible to maximize over all possible
subsets of N-grams.  Instead, we will assume that the N-grams
affect the relative entropy roughly independently, and compute 
$\dist{p}{p'}$ due to each individual N-gram.  We can then rank the
N-grams by their effect on the model entropy, and prune those that
increase relative entropy the least.

To choose pruning thresholds, it is helpful to look at a 
more intuitive interpretation of $\dist{p}{p'}$ in terms of 
{\em perplexity}, the average branching factor of the language model.
The perplexity of the original model (evaluated on the distribution
it embodies) is given by
\[
		\perpl = e^{- \sum_{h,w} p(h,w) \log p(w | h)} \quad ,
\]
whereas the perplexity of the pruned model on the original distribution is
\[
		\perpl' = e^{- \sum_{h,w} p(h,w) \log p'(w | h)}
\]
The relative change in model perplexity can now be expressed in
terms of relative entropy:
\[
	{\perpl' - \perpl \over \perpl} = e^{\dist{p}{p'}} - 1
\]

This suggests a simple thresholding algorithm for N-gram pruning:
\begin{enumerate}
\item
	Select a threshold $\theta$.
\item
	Compute the relative perplexity increase due to pruning each 
	N-gram individually.
\item
	Remove all N-grams that raise the perplexity by less than $\theta$,
	and recompute backoff weights.
\end{enumerate}

\paragraph{Relation to Other Work}

Our choice of relative entropy as an optimization criterion is by 
no means new.
Relative entropy minimization (sometimes in the guise of
likelihood maximization) is the basis of many model optimization
techniques proposed in the past, e.g., for text compression
\cite{Bell:90}, Markov model induction \cite{Stolcke:hmm-nips,Ron:94}.
Kneser \cite{Kneser:96} first suggested applying it to backoff N-gram models,
although, as shown in Section~\ref{sec:relation}, the heuristic pruning
algorithm of Seymore and Rosenfeld \cite{Seymore:96} amounts to
an approximate relative entropy minimization.
The algorithm described in the next section is novel in that it removes 
some of the approximations employed in previous approaches.
Specifically, the algorithm of \cite{Kneser:96} assumes that backoff weights
are unchanged by the pruning, and \cite{Seymore:96} does not consider
the effect that a changed backoff weight has on N-gram probabilities other
than the pruned one (this effect is discussed in more detail in
Section~\ref{sec:relation}).

The main approximation that remains in our algorithm is the greedy aspect:
we do not consider possible interactions between selected N-grams,
and prune based solely on relative entropy due to removing a single N-gram,
so as to avoid searching the exponential space of N-gram subsets.

\section{Computing Relative Entropy}
	\label{sec:computing}

We now show how the relative entropy $\dist{p}{p'}$ due to
pruning a single N-gram parameter can be computed exactly and efficiently.
Consider the effect of removing an N-gram consisting of history $h$ and
word $w$.  This entails two changes to the probability estimates.
\begin{itemize}
\item
	The backoff weight $\bow(h)$ associated with history $h$ is changed,
	affecting all backed-off estimates involving history $h$.
	We use the notation $\isbackoff(w_i,h)$ to denote this case, i.e.,
	that the original model does not contain an explicit N-gram estimate
	for $(w_i,h)$.
	Let $\bow(h)$ be the original backoff weight, and
	$\bow'(h)$ the backoff weight in the pruned model.
\item
	The explicit estimate $p(w|h)$ is replaced by a backoff estimate
\[
	p'(w | h) =  \bow'(h)  p(w|h') 
\]
	where $h'$ is the history obtained by dropping the first word in $h$.
\end{itemize}
All estimates not involving history $h$ remain unchanged, as do all estimates
for which $\isbackoff(w_i,h)$ is not true.

Substituting in (\ref{eq:delta}), we get
\begin{eqnarray}
\label{eq:delta-expand}
\dist{p}{p'} & = & - \mysum{w_i} p(w_i,h) [ \log p'(w_i|h) - \log p(w_i|h) ] \\
\nonumber
& = & \begin{array}[t]{l}
	- p(w,h) [ \log p'(w|h) - \log p(w|h) ] \\
	- \mysum{w_i: \isbackoff(w_i,h)} p(w_i,h) 
			[ \log p'(w_i|h) - \log p(w_i|h) ]
    \end{array} \\
\nonumber
& = & - p(h) \begin{array}[t]{l}
	\Bigl\{ p(w|h) [ \log p'(w|h) - \log p(w|h) ] \\
	+ \mysum{w_i: \isbackoff(w_i,h)} p(w_i|h)
				 [ \log p'(w_i|h) - \log p(w_i|h) ] \Bigr\}
	\end{array}
\end{eqnarray}
At first it seems as if computing $\dist{p}{p'}$ for a given N-gram requires a
summation over the vocabulary, something that would be infeasible 
for large vocabularies and/or models.
However, by plugging in the terms for the backed-off estimates, we see
that the sum can be factored so as to allow a more efficient computation.
\[
\begin{array}{l}
\dist{p}{p'} \\
= - p(h) \begin{array}[t]{l}
		\Bigl\{ p(w|h) \log p(w|h') + \log \bow'(h) - \log p(w|h) ] \\
   		+ \mysum{w_i: \isbackoff(w_i,h)} p(w_i|h) 
				 [ \log \bow'(h) - \log \bow(h) ] \Bigr\}
	\end{array} \\
= - p(h) \begin{array}[t]{l}
		\Bigl\{ p(w|h) [ \log p(w|h') + \log \bow'(h) - \log p(w|h) ]\\
		+ [ \log \bow'(h) - \log \bow(h) ]  
				\mysum{w_i: \isbackoff(w_i,h)} p(w_i|h) \Bigr\}
	\end{array}
\end{array}
\]
The sum in the last line represents the total probability mass 
given to backoff (the numerator for computing $\bow(h)$);  it needs to
be computed only once for each $h$, which is done efficiently by summing over
all {\em non-backoff} estimates:
\[
	\sum_{w_i: \isbackoff(w_i,h)} p(w_i|h) = 
		1 - \sum_{w_i: \neg \isbackoff(w_i,h)} p(w_i|h)
\]

The marginal history probabilities $p(h)$ are obtained by multiplying 
conditional probabilities $p(h_1) p(h_2|h_1) \ldots$.

Finally, we need to be able to compute the revised backoff weights
$\bow'(h)$ efficiently, i.e., in constant time per N-gram.  Recall that
\[
	\bow(h) = { 1 - \sum_{w_i: \neg \isbackoff(w_i,h)} p(w_i|h)
			\over
		    1 - \sum_{w_i: \neg \isbackoff(w_i,h)} p(w_i|h') }
\]
$\bow'(h)$ is obtained by dropping the term for the pruned N-gram
$(w,h)$ from the summation in both numerator and denominator.
Thus, we compute the original numerator and denominator once per history $h$,
and then add $p(w|h)$ and $p(w|h')$, respectively, to obtain $\bow'(h)$ for
each pruned $w$.

\section{Experiments}
	\label{sec:experiments}

We evaluated relative entropy-based language model pruning 
in the Broadcast News domain, using SRI's 1996 Hub4 evaluation system
\cite{Sankar:darpa97}.
N-best lists generated with a bigram language model were rescored with
various pruned versions of a large four-gram language model.%
\footnote{We used the 1996 system, partly due to time constraints,
partly because the 1997 system generated N-best lists using a large trigram
language model, which makes rescoring experiments with smaller language models
less meaningful.}

As noted in Section~\ref{sec:pruning}, the pruning algorithm is applicable
irrespective of the particular N-gram estimator used.  We used 
Good-Turing smoothing \cite{Good:53} throughout and did not investigate
possible interactions between smoothing methods and pruning.

\begin{table}
\begin{center}
\begin{tabular}{|c|r|r|r|c|c|}
\hline
$\theta$	& bigrams & trigrams	& 4-grams	& PP		& WER \\
\hline
0       & 11093357      & 14929826      & 3266900       & 163.0      & 32.6 \\
$10^{-9}$& 7751596       & 9634165       & 1938343       & 163.9       & 32.6 \\
$10^{-8}$& 3186359       & 3651747       & 687742        & 172.3      & 32.6 \\
$10^{-7}$& 829827        & 510646        & 62481	& 202.3      & 33.9 \\
\hline
0       & 11093357      & 14929826      & 0       	& 172.5      & 32.9 \\
\hline
\end{tabular}
\end{center}

\caption{\label{tab:hub4}
Perplexity (PP) and word error rate (WER) as a function of pruning 
threshold and language model sizes.}
\end{table}

Table~\ref{tab:hub4} shows model size, perplexity and word error results
as determined on the development test set, for various pruning thresholds.
The first and last rows of the table give the performance of
the full four-gram and the pure trigram model, respectively.
Note that perplexity here refers to the independent test set, not
to the training set perplexity that underlies the pruning criterion.

As shown, pruning is highly effective.
For $\theta= 10^{-8}$, we obtain a model that is 26\% the size of the original
model without degradation in recognition performance and less than 6\% 
perplexity increase.
Comparing the pruned four-gram model to the full trigram model, we see
that it is better to include non-redundant four-grams than to use
a much larger number of trigrams.  The pruned ($\theta = 10^{-8}$) four-gram
has the same perplexity and lower word error ($p < 0.07$) than the full
trigram.

\section{Comparison to Seymore and Rosenfeld's Approach}
	\label{sec:relation}

In \cite{Seymore:96}, Seymore and Rosenfeld proposed a different 
pruning scheme for backoff models (henceforth called the ``SR criterion,''
as opposed to the relative entropy, or ``RE criterion'').
In the SR approach, N-grams are
ranked by a weighted difference of the log probability estimate before
and after pruning,
\begin{equation}
	\label{eq:seymore}
	N(w,h) [ \log p(w|h) - \log p'(w|h) ]
\end{equation}
where $N(w,h)$ is the discounted frequency with which N-gram
$(w,h)$ was observed in training.
Comparing (\ref{eq:seymore}) with the expansion of $\dist{p}{p'}$ in
(\ref{eq:delta-expand}), we see that the two criteria are related.
First, we can assume that $N(w,h)$ is roughly proportional to
$p(w,h)$, so for ranking purposes the two are equivalent.
The difference of the log probabilities in (\ref{eq:seymore}) corresponds
to the same quantity in (\ref{eq:delta-expand}).
Thus, the major difference between the two approaches is that the SR criterion
does not include the effect on N-grams other than the one being considered,
namely, those due to changes in the backoff weight~$\bow(h)$.

To evaluate the effect of ignoring backed-off estimates in the pruning
criterion we compared the performance of the SR and the RE criterion
on the Broadcast News development test set, using the same N-best rescoring
system as described before.
To make the methods comparable we adopted Seymore and Rosenfeld's approach
of ranking the N-grams according to the criterion in question,
and to retain a specified number of N-grams from the top of the ranked list.
For the sake of simplicity we used a trigram-only version of the
Hub4 language model used earlier, and restricted pruning to trigrams.

We also verified that the discounted frequency $N(w,h)$ in (\ref{eq:seymore})
could be replaced with the model's N-gram probability $p(w, h)$ without 
changing the ranking significantly: over 99\% of
the chosen N-grams were the same.  This means the SR criterion can also be
based entirely on information in the model itself, making it more
convenient for model post-processing.

\begin{table}
\begin{center}
\begin{tabular}{|r|c|c|}
\hline
No.~Trigrams   & SR        & RE \\
\hline
1000           & 238.1     &  237.9 \\
10000          & 225.1     &  223.9 \\
100000         & 207.3     &  205.2 \\
1000000        & 186.4     &  184.7 \\
\hline
\end{tabular}
\end{center}
\caption{\label{tab:compare-ppl}
Comparison of Seymore and Rosenfeld (SR) and Relative Entropy (RE) pruning:
perplexities as a function of the number of trigrams.}
\end{table}

\begin{table}
\begin{center}
\begin{tabular}{|r|c|c|}
\hline
No.~Trigrams   & SR          & RE \\
\hline
0	     &   \multicolumn{2}{c|}{35.8} \\
1000         &   35.5         &           35.5 \\
10000        &   34.8         &           34.8 \\
100000       &   34.3         &           34.2 \\
1000000      &   33.2         &           33.1 \\
All	     &   \multicolumn{2}{c|}{32.9} \\
\hline
\end{tabular}
\end{center}
\caption{\label{tab:compare-wer}
Comparison of Seymore and Rosenfeld (SR) and Relative Entropy (RE) pruning:
word error rate as a function of the number of trigrams.}
\end{table}

Tables~\ref{tab:compare-ppl} and~\ref{tab:compare-wer} show model perplexity
and word error rates, respectively, for the two pruning methods as a
function of the number of trigrams in the model.
In terms of perplexity, we see a very small, albeit consistent, advantage for
the relative entropy method, as expected given the optimized criterion.
However, the difference is negligible when it comes to recognition performance,
where results are identical or differ only non-significantly.
We can thus conclude that, for practical purposes, the SR criterion is
a very good approximation to the RE criterion.

\begin{table}
\begin{center}
\begin{tabular}{|r|r|}
\hline
No.~Trigrams   & No. shared trigrams \\
\hline
1000    & 883 \\
10000   & 8721 \\
100000  & 85599 \\
1000000 & 852016 \\
\hline
\end{tabular}
\end{center}
\caption{\label{tab:compare-overlap}
Overlap of selected trigrams between SR and RE methods.}
\end{table}

Finally, we looked at the overlap of the N-grams chosen by the two criteria,
shown in Table~\ref{tab:compare-overlap}.
The percentage of common trigrams ranges from 88.3\% to 85.2\%, and seems to
decrease as the model size increases.  We can expect the most frequent
N-grams to be among those that are shared, making it no surprise that
both methods perform so similarly.

\section{Conclusions}
	\label{sec:conclusions}

We developed an algorithm for N-gram selection for backoff N-gram
language models, based on minimizing the relative entropy between the
full and the pruned model.
Experiments show that the algorithm is highly effective, eliminating
all but 26\% of the parameters in a Hub4 four-gram model without 
significantly affecting performance.
The pruning criterion of Seymore and Rosenfeld is seen to be an approximate 
version of the relative entropy criterion; empirically, the two 
methods perform about the same.

\let\secnums=n
\section{Acknowledgments}

This work was sponsored by DARPA through the Naval
Command and Control Ocean Surveillance Center under contract
N66001-94-C-6048.
I thank Roni Rosenfeld and Kristie Seymore for clarifications and 
discussions regarding their paper \cite{Seymore:96}.
Thanks also to Hermann Ney and Dietrich Klakow for pointing out similarities
to \cite{Kneser:96}.

\bibliographystyle{plain-shortnames}
\bibliography{all-short}

\section{Erratum}

The published paper had an error in the second equation for $\dist{p}{p'}$
in Section~\ref{sec:computing}.  In two instances, the quantity $\log \bow'(h)$
had been mistakenly typeset as $\log \bow(h')$.
Also, the information in reference \cite{Kneser:96} was incorrect.

\end{document}